\documentclass[10pt,journal,compsoc]{IEEEtran}
\ifCLASSOPTIONcompsoc
  \usepackage[nocompress]{cite}
\else
  \usepackage{cite}
\fi
\usepackage{graphicx}
\usepackage{subfigure}
\usepackage[cmex10]{amsmath}
\usepackage{mdwmath}
\usepackage{mdwtab}
\usepackage{booktabs}

\ifCLASSINFOpdf
\else
\fi
\begin{document}
%
\title{KGRGRL: A User's Permission Reasoning Method Based on Knowledge Graph Reward Guidance Reinforcement Learning}
%
%
%
%

\author{Lei~Zhang,
        Yu~Pan,
        Yi~Liu,
        Qibin~Zheng,
        and~Zhisong~Pan, 
\IEEEcompsocitemizethanks{\IEEEcompsocthanksitem Lei Zhang, Yu Pan and Zhisong Pan are with Command and Control Engineering College, Army Engineering University of PLA, Nanjing 210007, China. (E-mail: zhanglei@aeu.edu.cn, panyu@aeu.edu.cn, panzhisong@aeu.edu.cn).\protect
\IEEEcompsocthanksitem Yu Pan is with 31436 troop of PLA, Shenyang 110005, China. (E-mail: panyu@aeu.edu.cn).
\IEEEcompsocthanksitem Yi Liu and Qibin Zheng are with Academy of Military Science, Haidian, Beijing, 100091, China. (E-mail: albertliu20th@163.com, zqb1990@hotmail.com).
\IEEEcompsocthanksitem Both Lei Zhang and Yu Pan are contributed equally to this research.
\IEEEcompsocthanksitem Corresponding author: Zhisong Pan. E-mail:panzhisong@aeu.edu.cn.}
\thanks{Manuscript received xxxx, 2022; revised xxxx, 2022.}}

%
%

\markboth{IEEE Transcations on \LaTeX\ Class Files,~Vol.~14, No.~8, xxxx~2022}%
{Shell \MakeLowercase{\textit{et al.}}: Bare Demo of IEEEtran.cls for Computer Society Journals}
%



\IEEEtitleabstractindextext{%
\begin{abstract}
In general, multiple domain cyberspace security assessments can be implemented by reasoning user's permissions. However, while existing methods include some information from the physical and social domains, they do not provide a comprehensive representation of cyberspace. Existing reasoning methods are also based on expert-given rules, resulting in inefficiency and a low degree of intelligence. To address this challenge, we create a Knowledge Graph (KG) of multiple domain cyberspace in order to provide a standard semantic description of the multiple domain cyberspace. Following that, we proposed a user's permissions reasoning method based on reinforcement learning. All permissions in cyberspace are represented as nodes, and an agent is trained to find all permissions that user can have according to user's initial permissions and cyberspace KG. We set 10 reward setting rules based on the features of cyberspace KG in the reinforcement learning of reward information setting, so that the agent can better locate user's all permissions and avoid blindly finding user's permissions. The results of the experiments showed that the proposed method can successfully reason about user's permissions and increase the intelligence level of the user's permissions reasoning method. At the same time, the F1 value of the proposed method is 6\% greater than that of the Translating Embedding (TransE) method.
\end{abstract}

}

\maketitle

\IEEEdisplaynontitleabstractindextext

%
\IEEEpeerreviewmaketitle

\IEEEraisesectionheading{\section{Introduction}\label{sec:introduction}}

%
%
%
%

\IEEEPARstart{A}{rtificial} intelligence has long been a hot topic for identifying and mitigating cyberspace attacks. Because of the complexity and variety of cyberspace threats, collaborative modeling based on domain expertise has become the standard way for identifying them. However, while the component provides physical and social domain knowledge, the present approach lacks cyberspace to complete modeling, making it difficult to correctly infer the user's intent.

At this time, the physical domain, information domain, network domain and social domain in cyberspace are mostly described[1]. The physical domain is used to represent spatial information such as cities, regions, buildings, rooms, and so on. The typical information of cyberspace is the network domain. The social domain in cyberspace mostly refers to interpersonal relationships. For example, if an attacker and a company's cyberspace administrator were once classmates, the attacker will have an easier time obtaining cyberspace permissions than other attackers. The information domain, which primarily represents digital information such as a user name, password, key and information.

The main goal of the Knowledge Graph (KG) based description of multiple domain cyberspace is to extract semantic information from the design and configuration of multiple domain cyberspace and to determine the negative effects of event and configuration changes on the security state of cyberspace. The event described in cyberspace to the real quantity and entity relationship and the influence of the properties, as well as the change of entity and entity relationship's influence on the relationship between user permissions, can be defined as first-order logic corresponding reasoning rules, the event described in cyberspace to the real quantity and entity relationship and the influence of the features, and the change of entity and entity relationship's influence on the relationship between user's permissions, can all be defined as first-order logic corresponding reasoning rules. Therefore a formal description of the impact of cyberspace security events can be created. The primary goal of this method is to understand how different multiple domain cyberspace attacks interact. Existing reasoning methods, on the other hand, are founded on rules. First-order logic defines rules, and the corresponding reasoning rules must be given by experts. However, it has some limitations, such as a low level of intelligent. As a result, determining how to automatically extract user's final permissions from the KG of multiple domains in order to achieve intelligent reasoning of user's final permissions is an issue that merits more investigation.

This process can be divided into three steps: firstly, through theoretical analysis, hierarchical entity features are constructed according to top-down and bottom-up methods, mainly covering various entities in physical domain, information domain, network domain and social domain; Secondly, after determining various entities, it is necessary to sort out the relationships between entities, which are mainly divided into inclusion relationship, dependence relationship, dominance relationship, trust relationship and other types. Finally, after the entities and their relationships are determined, the multiple domain cyberspace KG is constructed to achieve a unified semantic description of multiple domain cyberspace.

By establishing good multiple domain cyberspace semantic information, can be defined first-order logic corresponding reasoning rules, the event described in cyberspace to the real quantity and entity relationship and the influence of the properties, as well as the change of entity and entity relationship's influence on the relationship between user permissions, found that the user is obtained than they should have permissions. So as to achieve the formal description of the impact of network security events. The fundamental purpose of this method is to grasp the interrelation between multiple domain cyberspace attacks. However, the existing reasoning methods are rule based reasoning. Rules are defined by first-order logic, and the corresponding reasoning rules need to be specified by experts. Therefore, it has certain limitations, that is the level of intelligence is low. Therefore, how to automatically reason user's final permissions from the multiple domain cyberspace KG, so as to realize intelligent reason of user's final permissions is a challenge.

To address this challenge, this paper uses the extraction of multiple domain cyberspace entities and relationship information for the construction of a multiple domain cyberspace KG, the KG to uniformly describe the multiple domain cyberspace. Based on multiple domain cyberspace KG, we proposed a user's permissions reasoning method based on reinforcement learning to realize the user's final permissions, learn how multiple domain cyberspace configuration and cyberspace entity relationships affect the ultimate permissions obtained by users, as well as intelligent reasoning to determine whether the user deserves the permissions. If user's final permissions is more than initial permissions. This indicates that cyberspace configuration has vulnerabilities, and the goal of the proposed method is to increase cyberspace security by further optimizing cyberspace setup.

The main contributions of this paper are as follows:
\begin{itemize}
	\item We are the first to consider reinforcement learning method for learning cyberspace KG relationships and to reason user's permissions;
	\item The multiple domain cyberspace is described uniformly by KG, and the security state of the existing multiple domain cyberspace is reasoning by the unified semantic information description, so as to effectively discover the relationship between multiple domain attacks;
	\item The proposed method realize the user's permissions reasoning method based on reinforcement learning, the method is abandoned the traditional pattern of experts write reasoning rules in advance, can let the model automatically learn reasoning rules, realize intelligent reasoning of user's permissions, have wider applicability and operability, provides a new train of thought for user's permissions intelligent reasoning. At the same time, the experimental results show that the proposed method has more advantages than the existing permissions reasoning methods.
\end{itemize}

In the next section, we introduced related works. We describe the proposed method in Section 3. We show experimental results in Section 4. Finally, we conclude in Section 5.

\section{Related Works}

KG is essentially a knowledge representation and a semantic network that reveals the relationships between entities \cite{2012Knowledge}. It belongs to the semantic network category and contains real world entities, relationships and events. KG is a knowledge representation, a semantic network that displays relationships between entities, it falls under the semantic network category, which comprises entities, relationships, and events in the actual world.

External objective facts are referred to as information, while external objective rules are inferred and summarized as knowledge. Knowledge is the ability to make connections between entities based on data. To put it another way, objects are made up of information that is represented as a subject predicate object. The KG operates in this way. Previously, the most common method for creating KG was to work from the top down, establishing the KG ontologies and data schema before adding things to the knowledge base. However, as a fundamental knowledge base, this creation approach must use an already organized knowledge base \cite{guo2021overview}.

\subsection{Construction of Cyberspace KG}
In general, the construction of knowledge atlas in cyberspace mainly includes ontology construction, information extraction \cite{zhao2020Knowledge} and knowledge storage \cite{wang2019research}. In reference \cite{undercoffer2003modeling}, an ontology is developed to model attacks and related entities, and the proposed ontology is only for attacks. In order to represent the concepts and entities related to the field of cyberspace security, reference \cite{joshi2013extracting} proposed a cyberspace security ontology based on the ontology of reference \cite{undercoffer2003modeling}. They extend the ontology to provide model relationships that capture the schema structure and security utilization concepts of the U.S. National Vulnerability Database. The ontology includes 11 entity types such as vulnerability, product, means and consequence. Reference \cite{2012BBBA} extends the ontology proposed in reference \cite{undercoffer2003modeling} and adds rules into the reasoning logic. Their ontology consists of three basic categories: methods, results, and goals.

There are two main approaches to information extraction: one is a knowledge-based engineering approach, which relies heavily on extraction rules but allows the system to deal with domain-specific information extraction problems. Most of the early information extraction systems are based on extraction rules. However, the main disadvantage is that it requires the participation of experts. Therefore, the extraction accuracy of extraction system is high. At present, many information extraction systems are based on knowledge engineering \cite{yang2018accurate}.

The second main method is based on machine learning \cite{7zhhe}. With the rise of artificial intelligence and machine learning, this method has become the mainstream method. The basic steps involve training an information extraction model with a large amount of training data, and then using the information extraction model to extract relevant information. The advantage of this method is that there is no need for experts to define rules in advance and the intelligence level is improved, but a large amount of training data is needed to achieve better experimental results. Reference \cite{2013Information} proposes a system that can identify relevant entities from unstructured text, which mainly solves the problems of network attacks and software vulnerabilities. Reference \cite{mulwad2011extracting} developed a framework for detecting and extracting vulnerability and attack information from network texts, and then trained a support vector machine to identify potential vulnerabilities. The classifier uses a standard one-word packet vector model. Once potential vulnerability descriptions are identified, the framework uses standard named entity recognition tools to extract security-related entities and concepts. The above methods all describe machine learning methods for automatically extracting relevant information from unstructured text. However, the method cannot accurately identify relevant entities until sufficient training data are obtained.

At the same time, the relationship between information units after information extraction is flat, lacking hierarchy and logic, and there are a lot of redundant or even wrong information fragments. Knowledge storage is the process of integrating knowledge from multiple knowledge bases to form a knowledge base. In this process, the main technologies include reference resolution, entity disambiguation and entity linking. Different knowledge base collect different key knowledge for the same entity, some knowledge base may focus on the description of its own some respects, some knowledge base may focus on the description of the entity and other entities, the relationship between the different knowledge base of knowledge storage of the real purpose is to describe the integration, in order to gain the complete entity description.

\subsection{Reasoning Method Based on Knowledge}
The practice of reasoning about unknown information based on current knowledge is known as knowledge reasoning. From individual to generic, by starting with known information and extracting new facts from it, or by drawing inferences from a big body of current knowledge. Symbol-based reasoning and statistics-based reasoning are two types of knowledge-based reasoning \cite{2004Domain}. Symbol-based reasoning is generally based on classical logic (first-order predicate logic or propositional logic) or versions of classical logic in artificial intelligence, such as default logic. Symbolic reasoning can discover logical conflicts between items as well as derive relationships between new entities from existing ones utilizing rules. Machine learning methods are commonly used in statistics-based reasoning approaches to discover new entity associations from KG.

The goal of knowledge reasoning is to figure out how instances and connections in KG are connected. To forecast the type of instances, reference \cite{paulheim2013type} provides an SDType approach that leverages a statistical distribution of characteristics connected by triples or predicates. The approach can be used to create a knowledge graph from a single data source, but it can't be used to reason knowledge across several data sets. Researcher approach has been presented as a tool for automatically inserting entities in the reference \cite{2012Automatic}, whereas linked related datasets \cite{kliegr2015linked} employ unique abstract data to extract instance types using specified schema. However, because this method relies on structured text data, it cannot be applied to other databases.

\section{Methods}

\subsection{Unified Description of Multiple Domain Cyberspace Semantic Information Based on KG}
In the construction of traditional knowledge atlas of cyberspace, most studies focus on network domain and information domain, which refers to network equipment and processing of network data traffic, and rarely involve physical domain, social domain and other fields. However, with the deepening of the research on cyberspace, academia and the industry have realized that cyberspace not only exists in the domain of cyberspace and digital domain, but also is affected by multiple domain behavior. To address this challenge, we propose a unified semantic information description method based on KG in multiple domain cyberspace. In reference \cite{2019MDC}, cyberspace should be integrated into physical domain, information domain, network domain, social domain and other domains, so it can be seen that cyberspace has the characteristics of multiple regions.

Firstly, the domain and scope of cyberspace entities are determined, and the relevant data of multi-domain cyberspace are integrated and standardized, providing a top-level model for constructing multiple domain cyberspace knowledge atlas. Second, after determining the scope of related entities, reuse of existing related entities is beneficial to improve construction efficiency.

(1) Entity information extraction

In this paper, we propose that there are $7$ types of entity information to be collected, namely, space entity, device entity, port entity, service entity, file entity, information entity and personnel entity.

Space entity belong to physical domains and are used to represent spatial information such as cities, campuses, buildings and rooms. At the same time, the physical domain also has device entities, namely physical entities. Device entity include not only network terminals, such as switches, routers, servers, and user terminals, but also related physical entities, such as keys and access cards. A port entity is located in a network domain and represents the physical ports of various network devices. It also contains not only physical ports but also virtual network ports. The service entity is also located in the network domain and represents the open services on the device, such as HTTP, FTP, and email services. File entity are located in the information domain and represent information entities and digital files, etc. Digital files can also represent digital files stored on terminals or servers. Information entity are located in information domains and are used to represent various types of information, such as user names, passwords, keys, and messages. Personnel entity represent the information of people involved in the network, including attackers, common users and administrators in the social domain.

(2) Semantic information extraction and construction

The KG pulls semantic information from multiple domain cyberspace configuration information to provide a unified semantic description of multiple domain cyberspace. This research focuses on the semantic information while extracting the semantic information of cyberspace setup. The lower-level semantic information facilitates the straightforward description of the semantic information of cyberspace configuration, although the model is huge and complex. The extracted fundamental semantic information in the unified description of multiple domain cyberspace based on KG mostly contains entity and relation. The following principles are used to generate the KG from the retrieved data:

\begin{itemize}
	\item Rule A: Group entities into a domain, and then group entities into the appropriate entity type
	\item Rule B: Extract relationships from cyberspace;
	\item Rule C: Connect related entities with wires and represent their relationships.
\end{itemize}

\subsection{User's Permissions Reasoning Method Based on Reinforcement Learning}
\begin{figure}[ht]
	\centering
	\includegraphics[width=\linewidth]{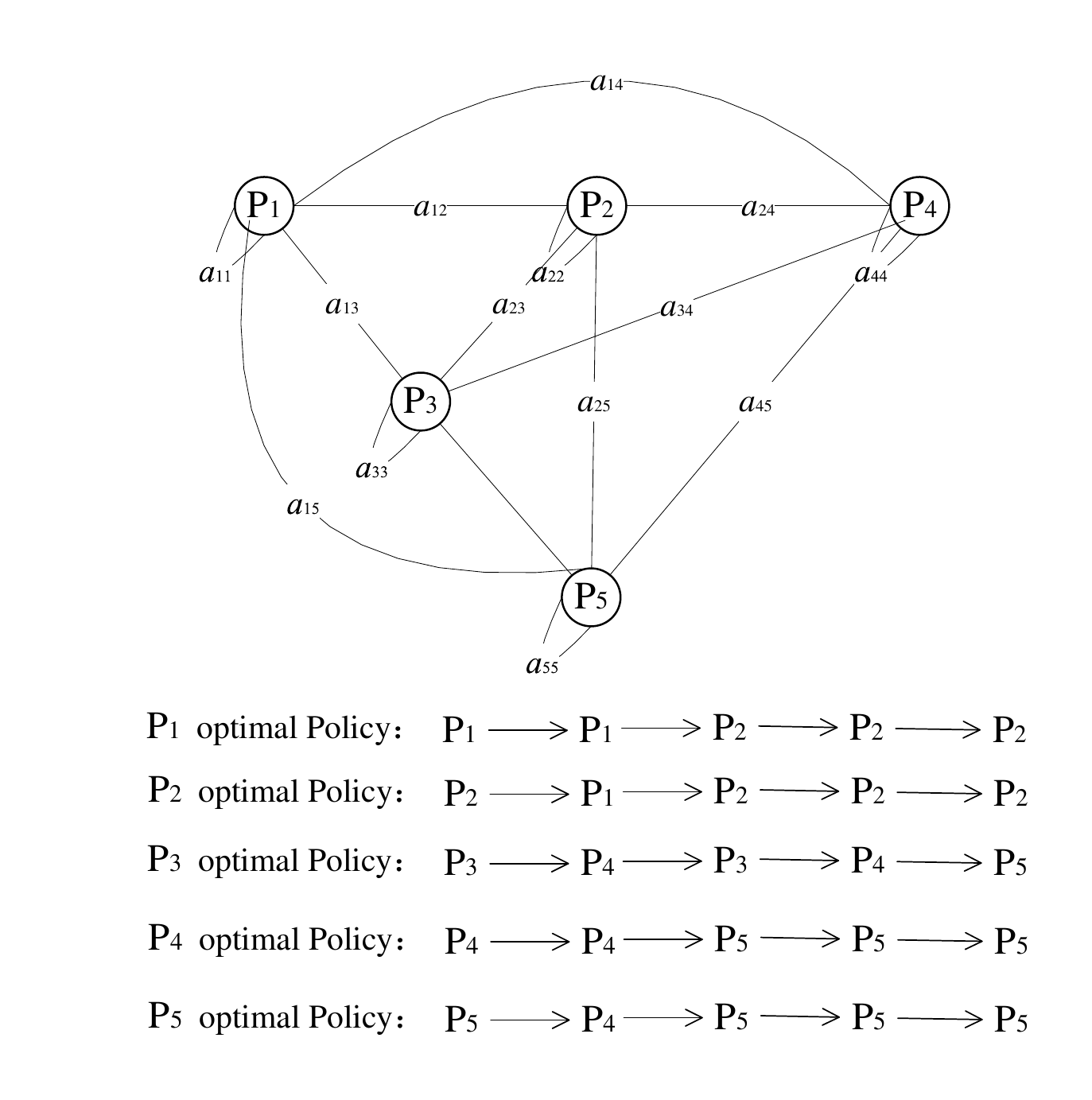}
	\caption{The Reasoning Methods}
	\label{Figure_1}
\end{figure}

We first introduce the basic elements of the RL framework in KG reasoning, including environment, state, action, and reward. As shown in Fig. \ref{Figure_1}, the input of the agent is a state composed of head and tail entities. The output is the next relationship predicted by the agent.

First, we should collect the following user permissions, including space access permission, device use permission, device control permission, port use permission, port control permission, service access permission, service control permission, file control permission, and information known permission.

Space access permission, device use permission, and device control permission belong to the user's permissions of the physical domain. Space access permission means that the user can enter the physical space, device use permission means that the user has the device use permission, and device control permission means that the user can control the device. The difference between device use permission and device control permission is that the former means that users can only use the device based on the configured status and cannot modify the configuration parameters of the device, while the latter means that users can modify the configuration parameters of the device, such as the firewall configuration and access control rights. Port access permission, port control permission, and service access permission. Service control permission refers to the network domain permission. Port control permission means that users can use the port to access the service, and port control permission means that users can change the port status or configuration. Service access means that the service request traffic will access the service, but it does not mean that the service can be used normally. The service control permission means that the user can pass the security authentication of the service and use the service normally. For example, the user can use the email service after being authenticated by the email server. File control permission and information known permission refer to information domain permissions. File control permission means that users can read, delete, and modify configuration files. Information known permission means that users know internal secret digital information such as passwords and keys.

Agent's search for permissions reasoning path is a process of trial and error. The application of reinforcement learning in KG reasoning is based on the assumption that as long as the agent can reach the tail entity $P_t$ from the head entity $P_1$ within a certain number of steps, we can regard this path as a potential reasoning path. DeepPath \cite{Xiong2017DeepPathAR} first introduced reinforcement learning into KG reasoning. The main task is to find the path from head to tail entity in the KG. In this method, the KG is sampled, the policy network is trained, and then the policy network is trained by the artificially designed reward function. We perform reasoning tasks through the agent, and every time the agent takes an action, a state transition will occur and a corresponding reward will be given to it. Our proposed method is consistent with the situational concept in DeepPath. But our method has different with DeepPath in the following details:

\begin{itemize}
	\item In DeepPath, the states is the entities and relations in a KG are naturally discrete atomic symbols. But in our proposed method, the states is the permissions that agent has. In Figure \ref{Figure_1}, if agent has $P_1$ permission, the state is agent located in $P_1$; If agent takes action $a_11$ to has permission $P_2$, the state is agent located in $P_2$.
	\item In DeepPath, the main task is to find the path from head to tail entity in the KG. In our proposed method, we limit the length of the path to $n$. In general, if agent has permission $P_t$, the permissions it's most likely to obtain are the permissions agent closest to finding. For example, in Figure \ref{Figure_1}, if user only have permission $P_1$, we use reinforcement learning to find $P_1$'s optimal policy (optimal path), and we set $n=5$, so we can reasoning if user has permission $P_1$, he will have permission $P_2$.
	\item In DeepPath, the reward will be given by specialist's experiences. In general, reward has a bigger contribute to the quality of the paths found by the agent. To encourage the agent to find optimal paths, we set the reward based on cyberspace KG, not by specialist's experiences. It is described in detail in the following. And this is our proposed method major innovation.
\end{itemize}

Next, we will cover each part of reinforcement learning in detail.

Environment: The entire user's permissions is considered to be an environment. This environment will remain unchanged throughout agent training. The environment also defines the interaction between agent and environment. That is, agent will change to a new state by interacting with the environment.

State: State encodes the position of agent in the environment with a vector of fixed length, that is, the position information of agent in the permissions graph.

Action: Our model treats each relationship type as an action. In permissions graph, permissions in graph is discrete atomic symbols. Due to the existing actual permissions, we simulate the symbolic atoms in all states. In our method, each state is the agent's position in the KG. The agent will then move from one entity to another when the operation is performed. The two are linked by action only taken by the agent.

We define the action space, where indicates whether the $i$ action is taken, 1 indicates that the action is taken, and 0 indicates no. The agent starts with the head entity, uses the policy network to take the most likely action in the current state, and further searches the path until it reaches the tail entity. The policy function maps the state vector to the probability distribution of all possible actions of $A$, namely:

\begin{equation}
	\pi_\theta(s,a)=p(a|s;\theta)
\end{equation}

Where, $\theta$ is calculated by the neural network and represents the parameters of the model.

Rewards: Rewards have always been the hardest part of reinforcement learning to set up. Most previous studies have assumed that the model only gets the final reward when it completes the relevant round, and that there is no single step reward. In this method, in addition to the final reward, we also set up different rewards for actions taken at different locations. We call this a KG reward guidance. 

The detail setting is following:

(1) If a user can enter one space and not be prevented by the security rules of that space and another space permission, he can have enter another space permissions. Therefore, the reward between the two permissions is setting $r=1$;

(2) If a user has access to a space permission, it has device use permission to all devices in that space. Therefore, the reward between the two permissions is setting $r=1$;

(3) If a user has the permission to use a device, it can have the permission to use all ports on the device. Therefore, the reward between the two permissions is setting $r=1$;

(4) If a user has the permission to use a port, he can have the permission using the port to access services accessible to the port. Therefore, the reward between the two permissions is setting $r=1$;

(5) If a user has the service reachability of a service permission, and he has the password for the service or the service does not have the password, he can obtain the permission control of the service. Therefore, the reward between the two permissions is setting $r=1$;

(6) If a user have control a service permission, he can have control the files from that service permission. Therefore, the reward between the two permissions is setting $r=1$;

(7) If a user have control a service permission, he can have the permission which controlling the information he gets from that service. Therefore, the reward between the two permissions is setting $r=1$;

(8) If a user have control a file permission, he can have control the information permission in that file. Therefore, the reward between the two permissions is setting $r=1$;

(9) If a user has access to a file permission and has the decryption key or the file is not encrypted, he can have the decrypted file permission. Therefore, the reward between the two permissions is setting $r=1$;

(10) If a user have control a service permission, he can control devices managed by that service. The device can be used and all ports on the device can be controlled. Therefore, the reward between permissions is setting $r=1$.
 
We set the reward information according to the above rules, and at the same time agent takes other actions, we set agent's reward $r=0$.


Policy network: We use a fully connected neural network to parameterize the policy function to map the state vector to the probability distribution of all possible actions. The neural network consists of two hidden layers, and the activation function uses ReLU function. The output layer uses SoftMax for normalization functions.

\section{Experiments}

In this section, in order to prove that multiple domain cyberspace can be uniformly described based on KG and the user's permissions can be reasoning based on KG, experiments are conducted to verify the correctness of the above methods. We set up a typical cyberspace environment and set it as experimental data sets. In this cyberspace environment, we not only simulate physical devices, physical connections and network services, but also include the physical space where they are, the location space of devices, digital files and information stored by different devices. From the perspective of user types, the network common users may have different permissions.

\subsection{Data Sets}

\begin{figure}[ht]
	\centering
	\includegraphics[width=\linewidth]{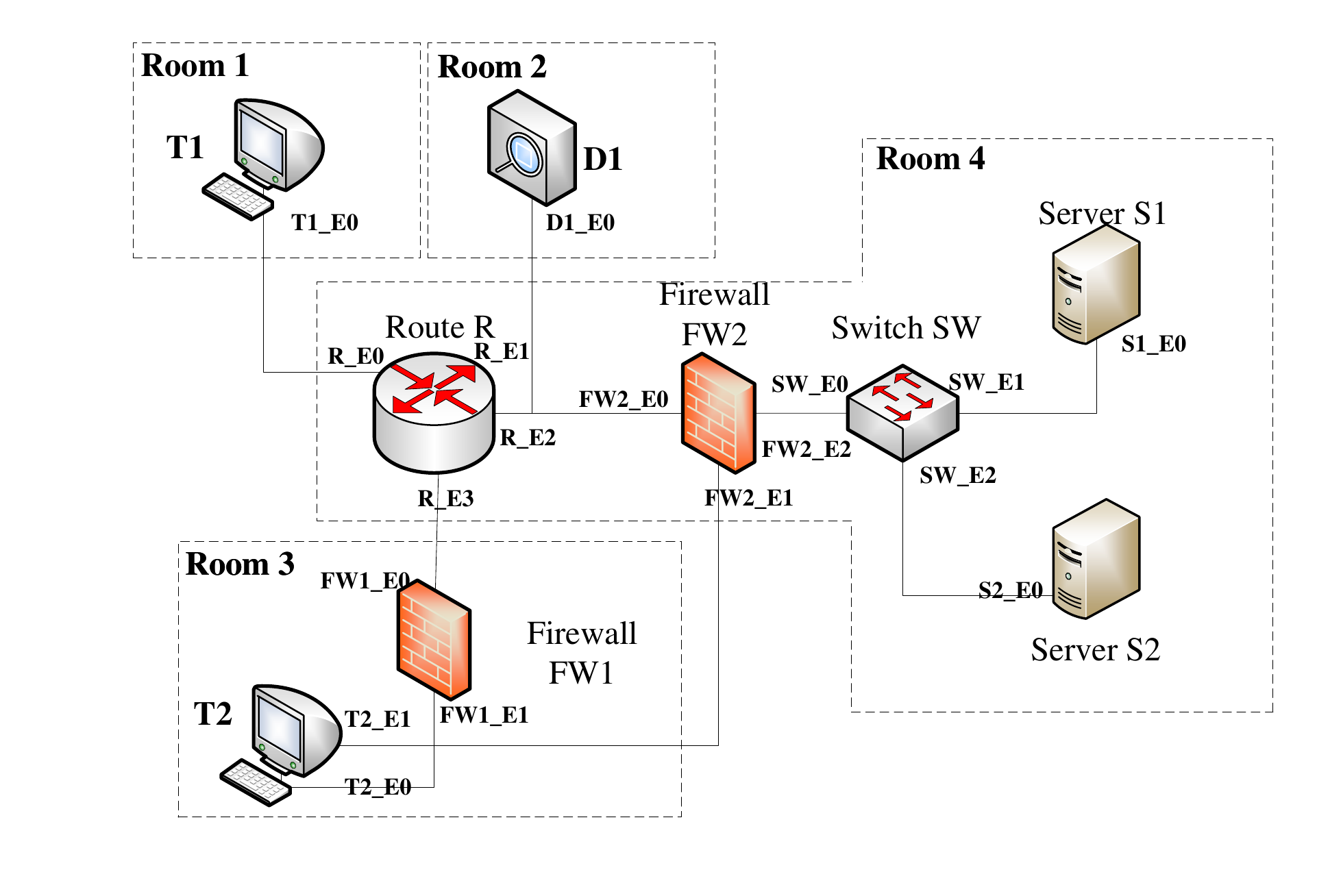}
	\caption{The Cyberspace Environment Topology}
	\label{Figure_2}
\end{figure}

The experimental data comes from the real environment of an enterprise network. In this environment, the outermost space is the entire external environment, representing a region, which can refer to campus, company, building, building, etc. Terminal T1 is in room 1, terminal T2 is in room 3, room 2 is in room D1, and room 4 is in room server. Two firewalls, FW1 in room 1, FW2 in room 4, router in room 4, switch in room 4, and two servers, S1 and S2 respectively, are located in room 4. Fig. \ref{Figure_2} shows the connections between devices.

\subsection{Baselines}

Method A: Path Ranking Algorithm (PRA) \cite{2016Knowledge}. This method makes use of the characteristics of graph structure, and makes use of the path relationship between entities to perform inference calculation, so as to directly estimate the relationship between two entity nodes. This method basically starts with a single entity node, where you are faced with two choices: either move to a randomly selected node or return to the starting node. The algorithm has only one parameter: restart probability $R$. Stability is achieved after iteration through countless random walks. The stabilized probability vector contains the score of all nodes in the network to the initial node, that is, the closeness between entity nodes. The node with the highest score is the entity node that can be reasoning.

Method B: Reasoning method Based on Rules (BOR) \cite{2019MDC}. Since there is no public permissions reasoning data sets in the previous research, we adopt the rule-based reasoning method based on \cite{2019MDC} to conduct user permissions reasoning for the training data sets and obtain the permissions that they can finally obtain. Although reasoning method based on rules has its limitations, the 
accuracy rate and recall rate of expert rule making are both 100$\%$. As described in \cite{2019MDC}, the accuracy of its rule-based user permissions reasoning method is 100$\%$.

Method C: TransE algorithm, whose core idea is to find a mapping function to transform each node in the graph into a low-dimensional dense embedded representation, requiring similar nodes in the graph to have the same distance in the low-dimensional space. The obtained representation vector can be used for downstream tasks, such as node classification, link prediction, visualization, etc. It is also essentially a knowledge-based approach.

Method D: User permissions reasoning based on KG reward guidance reinforcement learning (KGRGRL), which is proposed in this paper.

BOR is a rule-based reasoning method, in which experts specify rules, so its accuracy rate and recall rate are 100$\%$. The main advantage of our proposed method lies in its intelligence and refinement, and there is no data recognized by the academic circle for user authority reasoning. Therefore, the user authority reasoning of BOR method is used as training data of other methods to compare its effect.

\subsection{Evaluation Criteria}

Our judgment criteria for judging the correctness of reasoning are generally divided into two kinds: accuracy rate and recall rate. The definition formula is as follows:

\begin{equation}
	\begin{split}
	P=\frac{TP}{TP+FP} \\\\
  	R=\frac{TP}{TP+FN}
  	\end{split}
\end{equation}

$TP$ is the number of positive samples predicted as positive ones, $FP$ is the number of samples predicted as positive ones, and $FN$ is the number of samples predicted as negative ones.

\subsection{Results}

The device of our experiment is Intel X Power CPU, 64 GB memory, 2 pieces of Nvidia 2080 Ti GPU, and the operating system is Ubuntu 18.04. The experimental data sets are written by Python 3.6. We set up 5000 users, give them different initial permissions, and then get the corresponding end user permissions through BOR method.

\begin{table}[ht]
  \caption{Experiments Results}
  \centering
  \label{tab:freq}
  \begin{tabular}{cccl}
	    \toprule
	     Method&\emph{P}&\emph{R}&\emph{F1}\\
	    \midrule
	    PRA & 30.57$\%$ & 47.63$\%$ &37.24$\%$\\
	    TransE & 44.56$\%$& 56.89$\%$ &49.97$\%$\\
	    KGRGRL & \textbf{50.85$\%$} & \textbf{62.08$\%$} &\textbf{55.9$\%$}\\
	    BOR & 100$\%$ & 100$\%$ &100$\%$\\
	  \bottomrule
	\end{tabular}
	\end{table}

From the experimental results, the proposed method of accuracy and recall rate is not equal to zero, shows the proposed method can reason out the user's permissions, and permission to automated reasoning, and not rely on experts given rules, so our proposed method compared with BOR higher intelligent level, has a wider applicability. It can abandon the method of formulate reasoning rules by experts, and let the machine learn rules and features by means of machine learning, so as to achieve the purpose of user's permissions reasoning.

At the same time, compared with PRA method, the recall rate of the proposed method is greater than that PRA method, and the accuracy and recall rate of the proposed method are better than PRA method in the data set, indicating that the proposed method has a good effect and can improve the accuracy of existing reasoning methods to a certain extent.

Through the analysis of the experiment results, it was discovered that the user's permissions reasoning based on KG to reason out the user in all possible permissions ability under the current cyberspace configuration, can be based on existing knowledge, reasoning, the user may obtain permissions, abandon the traditional set of inference rules through the expert mode, solve the problem of its limited scope of application, and greatly improve the efficiency of reasoning.

\section{Conclusion}
This paper proposes a unified representation method and a user's permissions reasoning method based on reinforcement learning for semantic information in multiple domain cyberspace based on KG. Through knowledge mapping, entities and entity relations in multiple domain cyberspace can be described, so that entities in different domains of cyberspace can be described and expressed uniformly. Through to unify the existing multiple domain cyberspace semantic information description, thus effectively grasp the connection between the multiple domain cyberspace, based on this, realize the user's permissions based on knowledge reasoning, the reasoning method is to abandon the traditional expert writing mode of reasoning rules in advance, can let the machine learn automatic reasoning rules, implement the intelligent of the user's permissions. It has wider applicability and maneuverability and provides a new idea for intelligent reasoning of user's permissions reasoning. Through the experiment of simulation cyberspace environment, it is proved that this method can effectively deduce the user's final permissions under the current cyberspace configuration, and realize the intelligent reasoning of user's permissions. Therefore, the method proposed in this paper is feasible and effective.

This paper proposes a unified semantic description and reasoning method for multiple domain cyberspace based on KG, which can describe the information of different domains in cyberspace, so as to effectively describe and express the whole situation of cyberspace. But at the same time, our experiments are based on small-scale simulation cyberspace environment, and have not been effectively verified in large-scale real cyberspace environment. Next, we hope to be able to verify the correctness and validity of our proposed method in a large scale real cyberspace.


%



\ifCLASSOPTIONcompsoc
  \section*{Acknowledgments}
\else
  \section*{Acknowledgment}
\fi

The authors would like to thank this work was supported by the National Natural Science Foundation of China (No.62076251) and the National Key Research and Development Program of China (No.2017YFB0802800).

\ifCLASSOPTIONcaptionsoff
  \newpage
\fi



\bibliographystyle{IEEEtran}
\bibliography{IEEEabrv}
%



%

\begin{IEEEbiography}[{\includegraphics[width=1in,height=1.25in,clip,keepaspectratio]{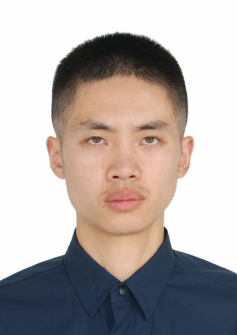}}]  
{Lei Zhang} was born in 1989 and received master's degree in Software Engineering from Army Engineering University of PLA in 2018. He is currently pursuing a Ph.D. in Computer Science and Technology in Command and Control Engineering College, Army Engineering University of PLA, Nanjing, China. His main research interests include reinforcement learning, machine learning, data mining and network security. \\
E-mail:zhanglei@aeu.edu.cn.
\end{IEEEbiography}
\begin{IEEEbiography}[{\includegraphics[width=1in,height=1.25in,clip,keepaspectratio]{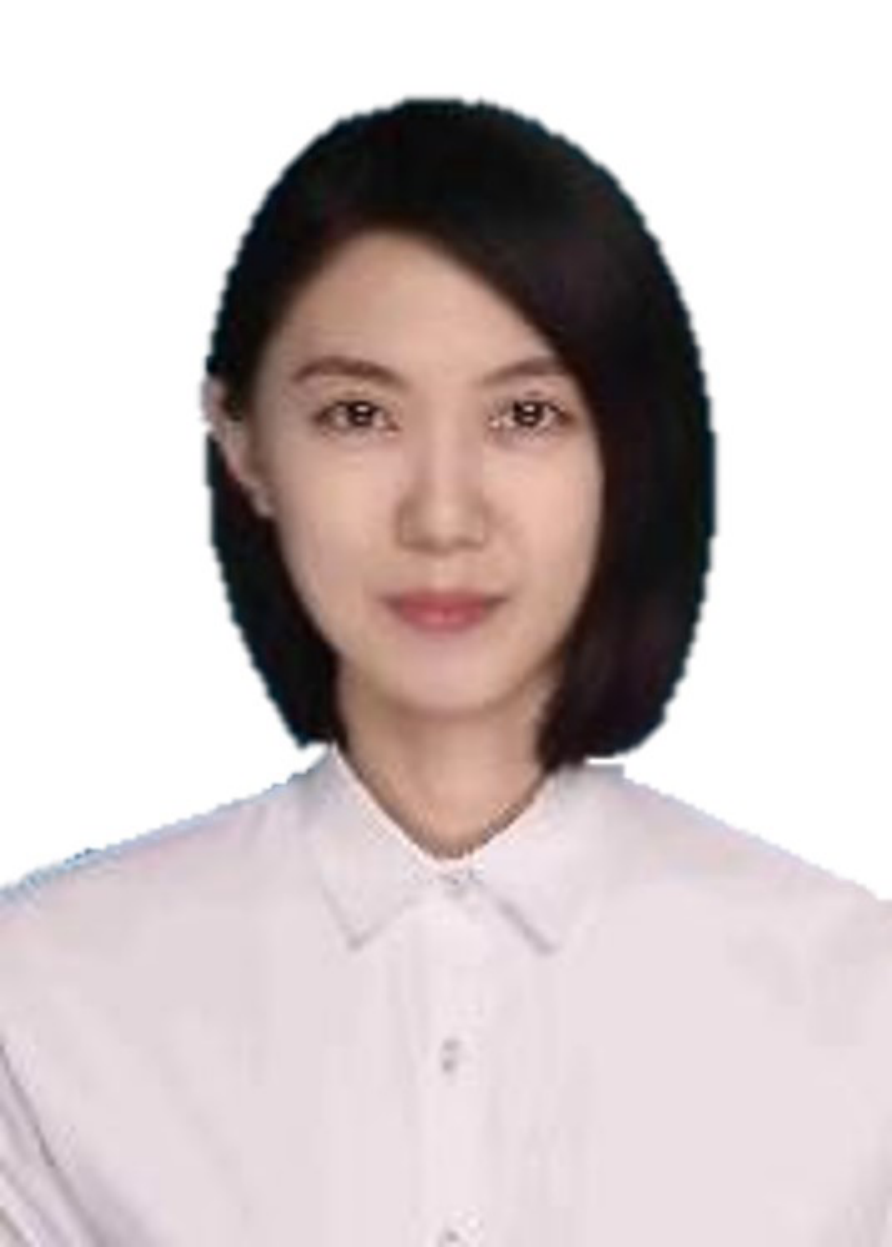}}] 
{Yu Pan} was born in 1990 and received the master's degree in Computer Science and Technology from Northeastern University, Shenyang, in 2015. She received her Ph.D. degree in Computer Science and Technology from Army Engineering University of PLA, Nanjing, in 2021. She is now an engineer in 31436 troop of PLA, Shenyang, Liaoning, 110005. Her main research interests include machine learning, data processing and mining in social networks.  \\
E-mail: panyu@aeu.edu.cn.
\end{IEEEbiography}
\begin{IEEEbiography}[{\includegraphics[width=1in,height=1.25in,clip,keepaspectratio]{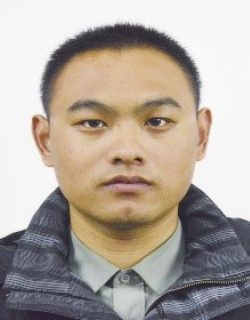}}] 
{Yi Liu} was born in 1990. He received the master's degree in Computer Science and Technology from the PLA University of Science and Technology and the Ph.D. degree in Software Engineering from Army Engineering University of PLA in 2018. He is now an assistant researcher in Academy of Military Science, 100091, Beijing, China. His main research interests include machine learning, evolutionary algorithms, and data quality. \\
E-mail: albertliu20th@163.com
\end{IEEEbiography}
\begin{IEEEbiography}[{\includegraphics[width=1in,height=1.25in,clip,keepaspectratio]{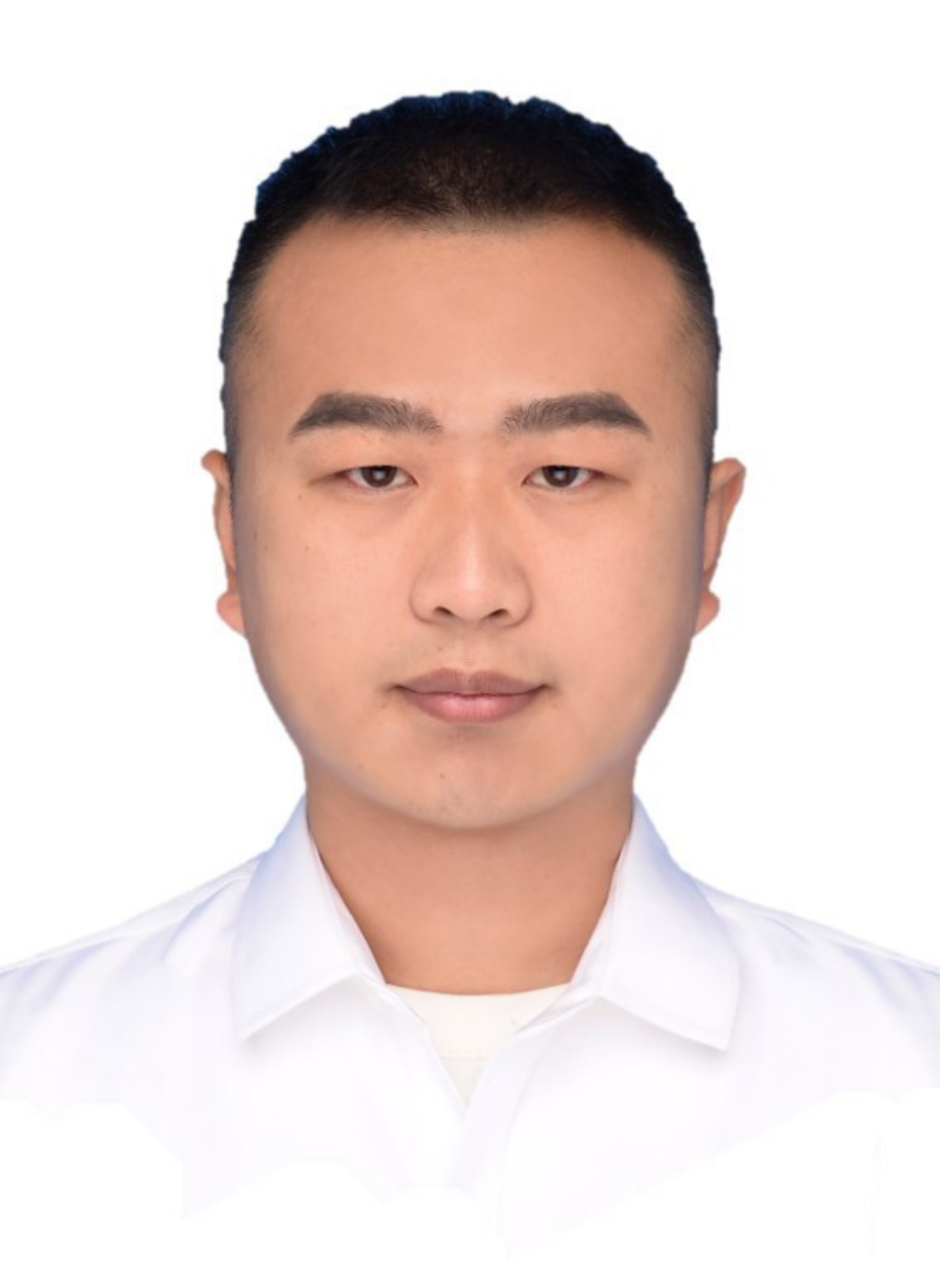}}] 
{Qibin Zheng} was born in 1990 and received the master's degree in Army Engineering University of PLA, Nanjing, in 2016, and has received the Ph.D. in Army Engineering University of PLA, Nanjing, in 2020. He is now an assistant researcher in Academy of Military Science, 100091, Beijing, China. His main research interests include data mining, machine learning and multi-modal data analysis.  \\
E-mail:zqb1990@hotmial.com.
\end{IEEEbiography}
\begin{IEEEbiography}[{\includegraphics[width=1in,height=1.25in,clip,keepaspectratio]{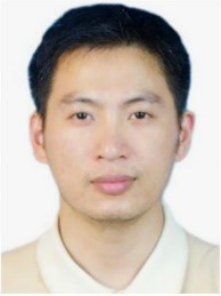}}] 
{Zhisong Pan} was born in 1973 and received Ph.D. degree from the Department of Computer Science and Engineering, Nanjing University of Aeronautics and Astronautics in 2003. Since July 2011, he has worked as a full professor and Ph.D. supervisor at Command and Control Engineering College, Army Engineering University of PLA, Nanjing, China. His current research interests include pattern recognition, machine learning and neural networks. 
\\ E-mail: panzhisong@aeu.edu.cn.
\end{IEEEbiography}







\end{document}